\documentclass[conference]{IEEEtran}
\renewcommand{\baselinestretch}{1.04}

\RequirePackage[normalem]{ulem} 
\RequirePackage{color}\definecolor{RED}{rgb}{1,0,0}\definecolor{BLUE}{rgb}{0,0,1} 

\RequirePackage{color}

\newcommand{\HRule}{\rule{\linewidth}{0.3mm}}
\usepackage{times}
\usepackage{graphicx}
\usepackage{amsmath}
\usepackage{amssymb}
\usepackage{endnotes}
\usepackage{stfloats}
\usepackage{cite}
\usepackage{enumerate}
\usepackage{color}
\usepackage{subfigure}
\usepackage{algorithm}
\usepackage{algorithmic}

\begin{document}

\title{\hspace*{-3pt}Hybrid Localization: A Low Cost, Low Complexity Approach Based on Wi-Fi and Odometry\vspace*{12pt}}

\author{
\IEEEauthorblockN{\emph{Letizia Moro,~\IEEEmembership{Student Member, IEEE,}}}
\IEEEauthorblockA{
\emph{Timberline High School, Boise, Idaho, USA}\\
\emph{Email: letizia.moro@gmail.com}}
\and
\IEEEauthorblockN{\emph{Hani Mehrpouyan,~\IEEEmembership{Member, IEEE,}}}
\IEEEauthorblockA{\emph{Department of Electrical and}\\
\emph{Computer Engineering,}\\
\emph{Boise State University, Boise, Idaho, USA}\\
\emph{Email: hani.mehr@ieee.org}
}
}

\maketitle
\thispagestyle{empty}
{\let\thefootnote\relax\footnotetext{{This research was supported by the NASA University Leadership Initiative Grant.\vspace{-0pt}}} }

\begin{abstract}
Localization in indoor environments is essential to further support automation in a wide array of scenarios. Moreover, direction-of-arrival knowledge is essential to supporting high speed millimeter-wave (mmWave) links in indoor environments, since most mmWave links are of a line-of-sight nature to combat the high pathloss in this band. Accurate wireless localization in indoor environments, however, has proved a challenging task due to multi-path fading. Additionally, due to the effects of multi-path fading, methods such as trilateration alone do not result in accurate localization. As such, in this paper we propose to combine the knowledge of wireless localization methods with that of odometry sensors to track the location of a mobile robot. This paper presents significant real-world localization measurement results for both Wi-Fi and odometry in diverse environments at the Boise State University campus. Using these results, we devise an algorithm to combine data from both odometry and wireless localization. This algorithm is shown in hardware testing to reduce the localization error for a mobile robot. 
\end{abstract}

\vspace*{8pt}

\vspace*{4pt}
\section{Introduction}

Robot positioning has been a major challenge in the automation industry. If accurate robot positioning can be achieved repeatedly and with economical feasibility, it would advance the automation industry a step closer to achieving a state of higher autonomy for multiple applications. Such wide deployment of robots, however, may be difficult to achieve without establishing accurate localization in indoor environments~\cite{borenstein1997mobile,biswas2010wifi}. This localization information is needed to effectively navigate the environment and to avoid any obstacles and collisions. Moreover, the next generation of wireless networks that is expected to more widely support Internet of Things (IoT) is expected to widely use millimeter-wave (mmWave) links to support indoor wireless links~\cite{wang2018iot}. Such links could use this localization information in conjuction with highly directional antennas to overcome significant pathloss at mmWave frequencies~\cite{mehrpouyan2014improving}. 

Given the above, the topic of robot localization has been an active area of research in which various approaches have been used. These methodologies can be summarized under two categories: 1) relative positioning approaches\footnote{Past localization measurement values impact current positioning accuracy, i.e. memory based.}, e.g. odometry, inertial navigation, etc.; and 2) absolute positioning approaches\footnote{Past localization measurement values have no impact on the current localization values, i.e. memoryless.}, e.g., magnetic compasses, active beacons, landmark navigation, etc.~\cite{borenstein1997mobile}. Given the focus of this paper, we summarize the important prior work related to active beaconing and odometry. 

\begin{figure*}[!hb]
\HRule
\begin{align}
	\label{eq:trilateration1}
	x=\frac{(d^2_1-d^2_2-x_A^2+x_B^2-y_A^2+y_B^2)(-2y_B+2y_C)-(d^2_2-d^2_3-x_B^2+x_C^2-y_B^2+y_C^2)(-2y_A+2y_B)}{(-2y_B+2y_C)(-2x_A+2x_B)-(-2y_A+2y_B)(-2x_B+2x_C)}\\
	\label{eq:trilateration2}
	y=\frac{(d^2_1-d^2_2-x_A^2+x_B^2-y_A^2+y_B^2)(-2x_B+2x_C)-(d^2_2-d^2_3-x_B^2+x_C^2-y_B^2+y_C^2)(-2x_A+2x_B)}{(-2y_A+2y_B)(-2x_B+2x_C)-(-2y_B+2y_C)(-2x_A+2x_B)}
\end{align}
\vspace{-10pt}
\end{figure*}
\setcounter{equation}{2}

\subsection{Related work}
As the title of this paper suggests, Wi-Fi signaling is used as an active beacon method to achieve indoor localization~\cite{honkavirta2009comparative, maxim2008trilateration}. This choice is motivated by the widespread use of Wi-Fi access points in many indoor environments. This significantly reduces the cost associated with indoor localization as beacon points are readily available, powered, and in use. 

The main approach in an active beaconing method for indoor Wi-Fi localization is through Wi-Fi fingerprinting~\cite{honkavirta2009comparative,koski2010positioning,koski2010indoor,
addesso2010adaptive,wu2017mitigating,tao2018novel,guo2018accurate,
sun2018augmentation,wang2017csi}. Wi-Fi fingerprinting, however, suffers from two main shortcomings: 1) The process of associating the Wi-Fi signal strength to every given location within the environment of interest, which can be complex and time consuming~\cite{tao2018novel}: 2) The presence of large errors due to obstacles, human shadowing, and many other factors that constantly influence and vary the Wi-Fi signal strength at various locations within an indoor environment~\cite{wu2017mitigating}. In other words, even small changes in the environment, e.g., introduction of new objects or individuals, could change the fingerprinting map and require retraining. 

To address the above issues, the work in~\cite{wu2017mitigating} focuses on reducing the error associated with Wi-Fi fingerprinting, by attempting to create specific methods to determine outliers and quantify the error associated with each access point (AP). Although the results in~\cite{wu2017mitigating} indicate that this method may be effective for smartphones, the proposed method does not address the complexity and overhead associated with the Wi-Fi fingerprinting process. On the other hand, the work in~\cite{tao2018novel} attempts to reduce the Wi-Fi fingerprinting overhead by using an automation algorithm which requires the device or user to monitor traffic from various APs and use this traffic information with the measured signal strength to construct a Wi-Fi fingerprinting map. Although the results are promising, little information is provided on how much reduction in overhead is achieved by using the algorithm described in~\cite{tao2018novel}. Moreover, the approach in~\cite{tao2018novel} can be power hungry since it requires the constant monitoring of data from a large number of access points. In fact, due to this overhead and complexity, new approaches are focusing on applying deep learning methodologies to reduce the overhead associated with Wi-Fi fingerprinting while also increasing accuracy~\cite{wang2017csi}. However, as stated above, these methods require constant retraining which is a time consuming process.

Given the above challenges, in this paper we propose to use Wi-Fi signals and trilateration to reduce and/or completely eliminate the need for Wi-Fi fingerprinting. Clearly, the application of trilateration in indoor settings will result in large errors. As such, we propose to use sensor fusion. This can be described as combining odometry results obtained through a set of sensors independent from Wi-Fi, in order to enhance localization accuracy for a mobile robot.

\begin{figure*}[!ht]
  \begin{minipage}[t]{0.48\textwidth}
     \vspace{-128pt}
     \includegraphics[scale=0.062]{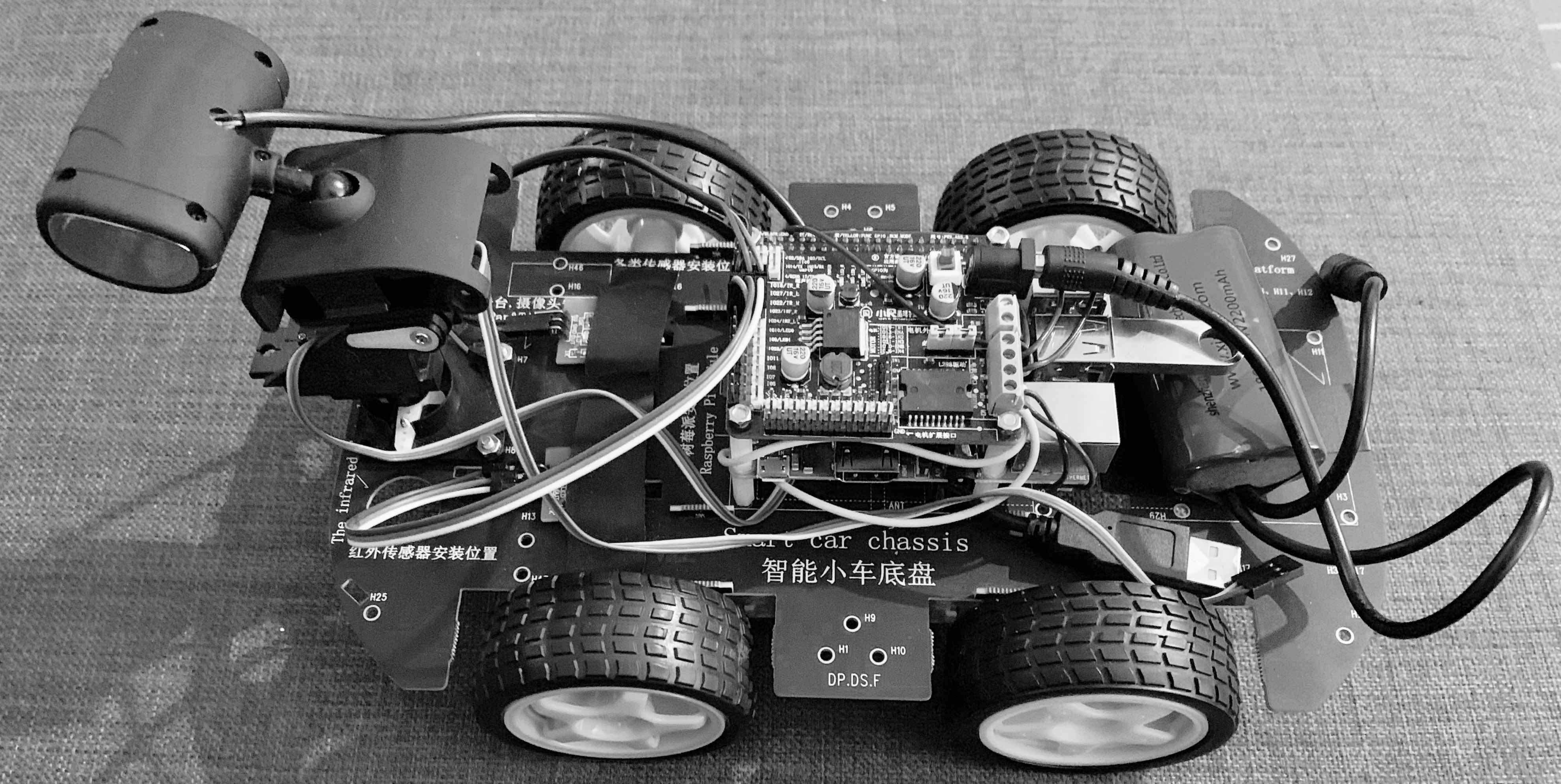}
    \vspace*{+5pt}
    \caption{Mobile robot used for measurments.} \label{fig_robot}
  \end{minipage}
  \begin{minipage}[t]{0.48\textwidth}
    \hspace{20pt}
    \includegraphics[scale=0.40]{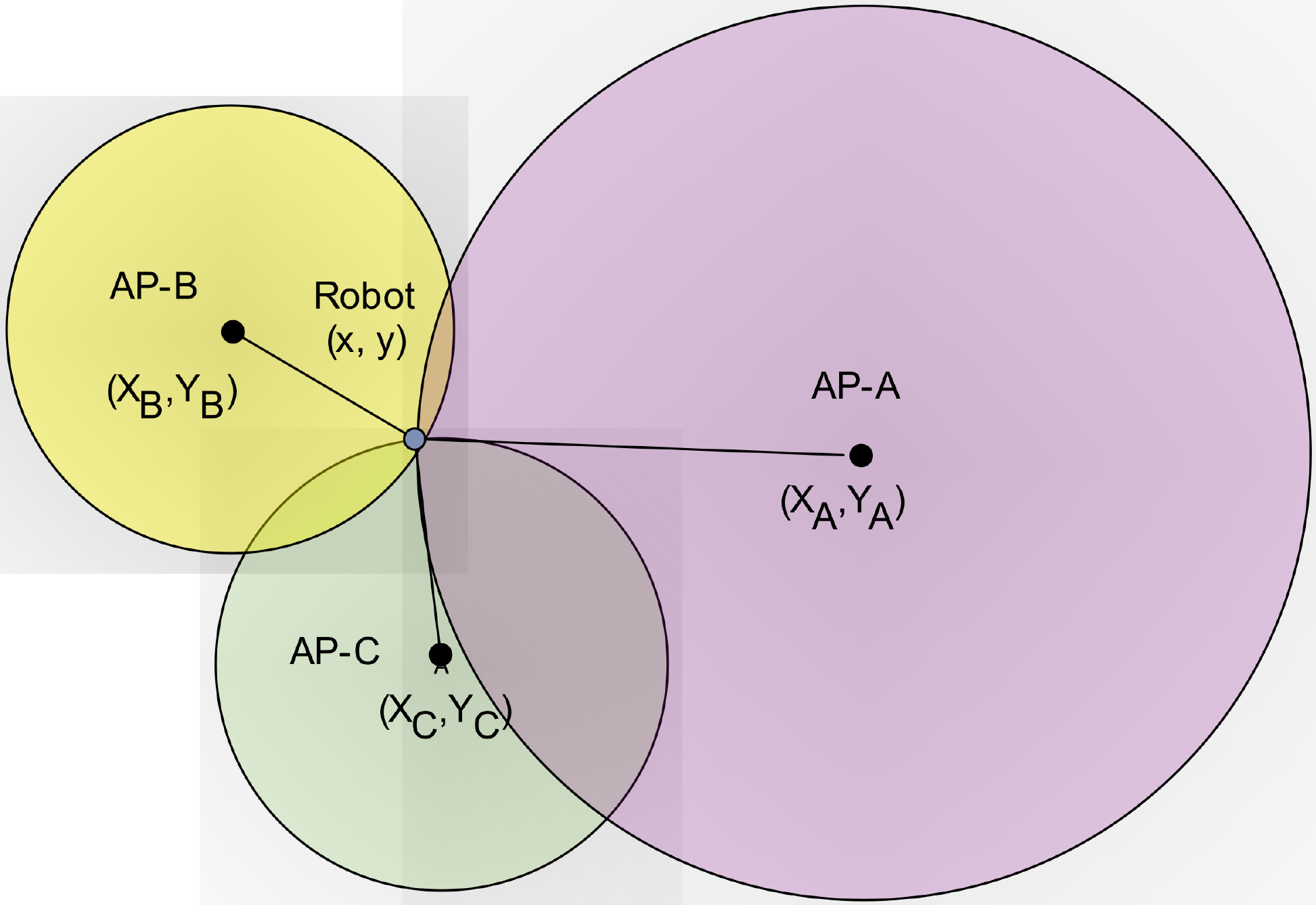}
   \vspace*{-0pt}
   \caption{Localization setup for trilateration.} \label{fig_trilateration}
  \end{minipage}
\end{figure*}

\subsection{Contributions}

In this paper, we propose a hybrid localization solution that is low cost, low complexity, and utilizes Wi-Fi trilateration and odometry to localize a robot in an indoor environment. Our real-world experimental results support that the combination of trilateration and odometry can result in accurate localization in various indoor environments. The specific contributions of this paper can be summarized as: 

\begin{itemize}
\item Unlike prior work that is mainly based on Wi-Fi fingerprinting, the proposed approach requires little overhead to start Wi-Fi localization in an indoor environment. The main information that is needed is the location of the three access points used for trilateration in the environment of interest. Given that these access points are placed in fixed, known locations, this information is readily available.
\item As stated in Wi-Fi fingerprinting literature~\cite{honkavirta2009comparative,koski2010positioning,koski2010indoor,
addesso2010adaptive,wu2017mitigating,tao2018novel,guo2018accurate,
sun2018augmentation,wang2017csi}, due to human shadowing, obstacles, and multi path fading, localization data extracted with Wi-Fi signaling could be affected by large outliers and errors. As such, we propose to fuse the data from odometry localization with that of Wi-Fi trilateration. Our real-world measurements show that odometry localization retains significant memory in the process, while Wi-Fi trilateration localization has a memoryless characteristic. Hence, the combination of both approaches results in accurate indoor localization. 

\item We carry out significant real-world measurements using an off-the-shelf mobile robot in various environments such as an office, a hallway, and a large arena- the Taco Bell Arena- at the Boise State University Campus to verify the potential and drawbacks of the proposed algorithm. 
\end{itemize}

This paper is organized as follows: Section II formulates the system setup for the proposed positioning algorithm. Section III outlines the proposed hybrid localization algorithm. Section IV presents the results of our real-world measurements, and Section V concludes the paper and proposes future research directions.

\vspace*{4pt}
\section{System Model}\label{section_system_model}
In this section, we present the system setup and the models used for both Wi-Fi and odometry. We focus on the $2.4$ GHz band of the IEEE 802.11 for localization. Fig.~\ref{fig_robot} presents the robotic vehicle used for testing and measurement. As shown in Fig.~\ref{fig_robot}, the robot is powered by $4$ independent electric motors at each wheel. A Raspberry Pi 3 Model B, with limited processing power, is used to run the algorithm proposed here. This supports our claim with regard to the low complexity nature of the proposed algorithm. The on-board Wi-Fi chip on the Raspberry Pi is used for wireless beaconing. Each wheel is equipped with low-cost infrared sensors that allow for precise monitoring of the rotation of each wheel. More specifically, the proposed setup allows us to detect $1/20$ or $1.02$ cm of a wheel rotation for each wheel.

Fig.~\ref{fig_trilateration} outlines the localization framework in this paper. As noted in this figure, we focus on two-dimensional localization in which the $x$ and $y$ coordinates of the robot are of interest. Moreover,~\eqref{eq:trilateration1} and \eqref{eq:trilateration2} are provided to obtain the coordinates of $x$ and $y$ via trilateration, respectively.

\subsection{Wireless Channel Model}

As demonstrated in~\eqref{eq:trilateration1} and \eqref{eq:trilateration2}, the distances $d_1$, $d_2$, and $d_3$ need to be measured in order to find the coordinates of the robot. Here, we use the close-in free space reference distance path loss model to obtain these values. The close-in free space reference distance path loss model~\cite{rappaport2014millimeter,maccartney2014omnidirectional,rappaport2015wideband} is represented by the path loss exponent, $n$ and $\sigma$ is given by
\begin{align}
	\label{eq1}
	\text{PL}(d)[\text{dB}] = \text{PL}(d_{o}) &+ 10 n \cdot \log_{10} ( \frac{d}{d_{o}} )
	\nonumber
	\\
	&+ \chi_{\sigma}, \;\;\;\;\;\;\;\;\;\;\;\;\;\;\;\;\;\;\;\;\;\textrm{for} \quad d \geq d_{o},
\end{align}
where, $d_{o} $ is the close-in free space reference distance, $\text{PL}(d_{o})$ is the close-in free space path loss in dB, and is given by \eqref{eq2} which is a function of wavelength or frequency.
\begin{equation}
	\label{eq2}
	\text{PL}(d_{o}) = 20\log_{10} \frac{4\pi do}{\lambda}
\end{equation}
Here, $\chi_{\sigma}$ is a normal random variable with $0$ dB mean and standard variation $\sigma$ \cite{rappaport2014millimeter}. The parameters of this model are well-known for various environments at the $2.4$ GHz and are readily available in many references, e.g.~\cite{goldsmith2005wireless}. 


\subsection{Odometry Model}
\begin{figure}[!b]
\begin{center}
    \scalebox{0.42}{\includegraphics {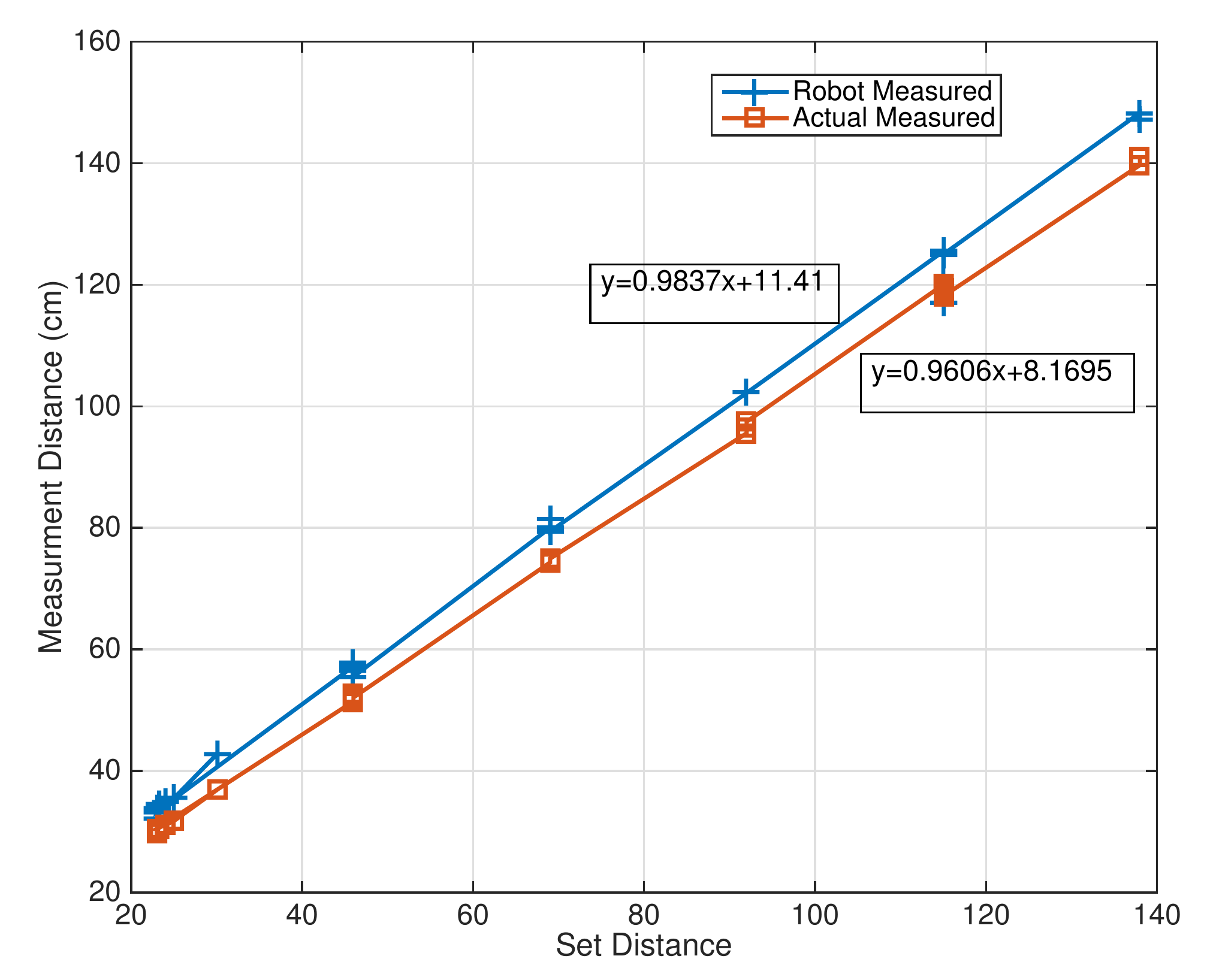}}
    \caption{The measurement results for actual traveled distance by the robot vs. the set values in the program. The equations for the trend-line are also indicated in this figure. }
    \label{fig_odo}
\end{center}
\end{figure}

Prior work in the field of odometry has shown that a linear model can consistently be used to achieve localization via odometry~\cite{borenstein1997mobile}. Our measurement results, as illustrated in Fig.~\ref{fig_odo}, also confirm this fact and indicate that the odometry results for the robot also follows a linear model. Moreover, for directional tracking, we monitor the rotation of each wheel and obtain specific values on the angular rotation of the robot due to the relative motion of each wheel. This allows us to track both the direction and distance the robot has traveled. 

\vspace*{4pt}
\section{Localization Algorithm}\label{sec_ALG}
In this section, we outline the proposed low complexity localization algorithm.

\subsection{Preliminary Information} 
\label{Prelim}
\begin{figure}
\begin{center}
\vspace{+0pt}
\scalebox{0.30}{\includegraphics {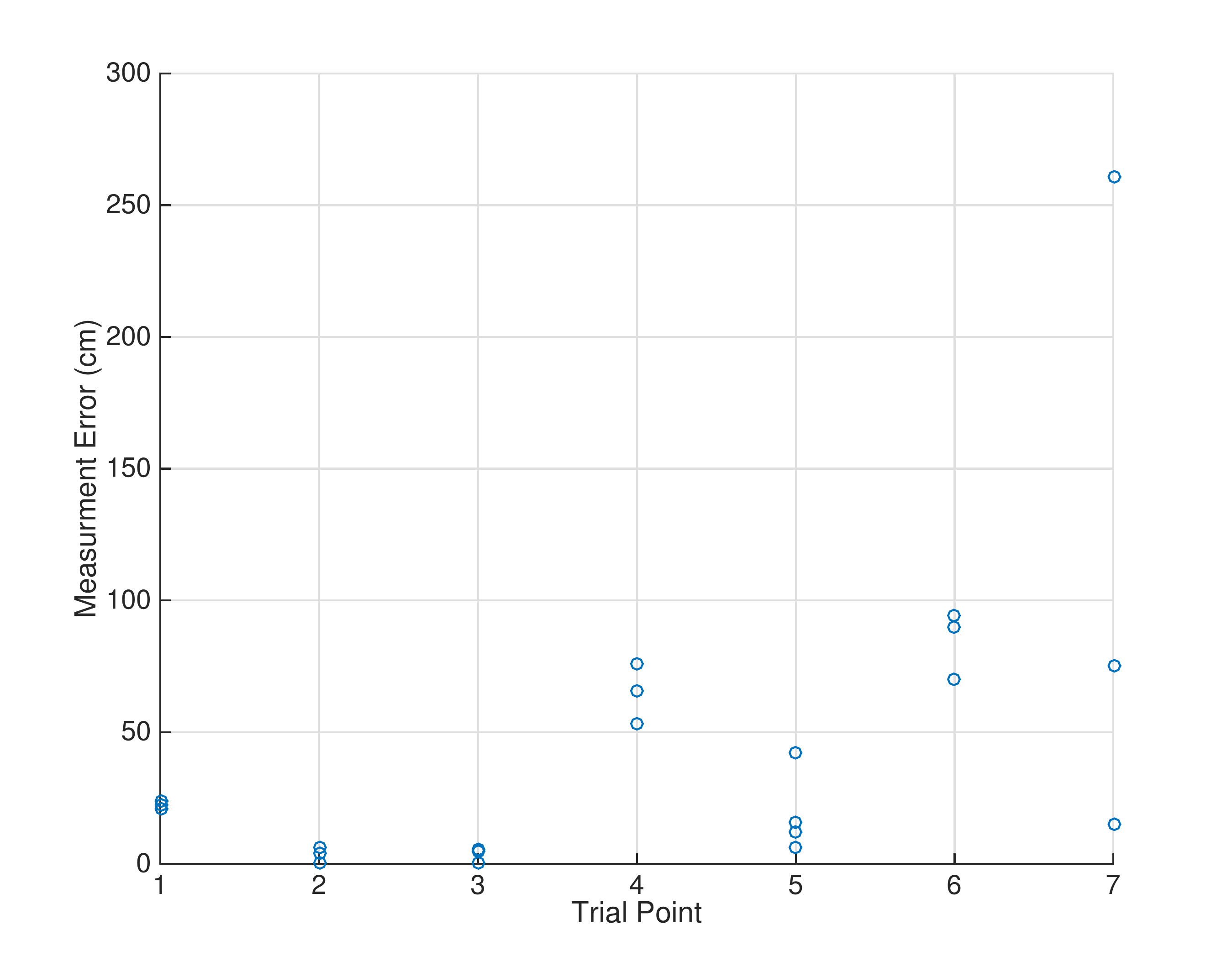}}
\caption{The results regarding the error measured from the Wi-Fi positioning  method at each trial point in the office setting. }
 \label{fig_Wi-Fi}
\end{center}
\end{figure}

Fig. ~\ref{fig_Wi-Fi} presents the error in Wi-Fi localization in the office environment of Fig.~\ref{fig.OFFICE}. As shown in this figure, at specific locations, due to the presence of a line-of-sight (LoS) link from the APs to the robot, Wi-Fi localization is accurate. However, due to obstacles, reflection, and human shadowing in an indoor environment, the Wi-Fi signal strength can vary significantly at various locations that may be geographically close~\cite{goldsmith2005wireless}. This variation is represented by the log normal shadowing coefficient, $\chi_{\sigma}$, in \eqref{eq1}. As such, the Wi-Fi localization algorithm can result in large outliers within its dataset.

Unlike Wi-Fi localization, which is an absolute method, odometry localization is a relative method. As such, although the errors for odometry localization are small, over time, they can accumulate. This is observed in our experimental results in Section~\ref{sec_exp} and is also confirmed in \cite{borenstein1997mobile}. Accordingly, we can use odometry to determine the outliers in Wi-Fi localization since, in our results, the error in odometry localization tends to grow slowly, while we can use Wi-Fi localization to recalibrate odometry measurements. The latter follows from the fact that Wi-Fi localization results in accurate localization between the outliers, as illustrated in Fig.~\ref{fig_Wi-Fi}.

\begin{algorithm}[t]
\caption{Proposed Hybrid Localization Algorithm}
\label{al_main}
\begin{algorithmic} 
\STATE Initialize $\omega_{\text{odo}}$, $\omega_{\text{Wi-Fi}}$, odoCounter, $x_{\text{odo}}$, $y_{\text{odo}}$, $x_{\text{Wi-Fi}}$, $ y_{\text{Wi-Fi}}$
\STATE Initialize $(x_A, y_A)$, $(x_B, y_B)$, $(x_B, y_B)$
\STATE Obtain ($x_{\text{odo}}$, $y_{\text{odo}}$) by monitoring rotation of each wheel and using the previous location of the robot
\STATE Obtain ($x_{\text{Wi-Fi}}$, $y_{\text{Wi-Fi}}$) by measuring signal strength and using \eqref{eq:trilateration1}, \eqref{eq:trilateration2}, and \eqref{eq1}
\STATE Obtain $\omega_{\text{odo}}$, $\omega_{\text{Wi-Fi}}$ by calling Dynamic Weight-Allocation Algorithm
\STATE Calculate $(x,y)$ of robot by using \eqref{eq:WeightAlg}
\end{algorithmic}
\end{algorithm}

\subsection{Proposed Algorithm}
The information outlined in Section~\ref{Prelim} illustrates the need and opportunity to create an algorithm to successfully combine each of the data sets from odometry and Wi-Fi. Here, we provide a modified weighted-average algorithm for fusing the localization results from Wi-Fi and odometry. 

We use the following mathematical formulation for fusing the Wi-Fi and odometry localization results if the Wi-Fi localization results are not deemed to be outliers.
\begin{align}\label{eq:WeightAlg}
	 x &= x_{\text{odo}}  \omega_{\text{odo}} + x_{\text{Wi-Fi}}  \omega_{\text{Wi-Fi}} \nonumber \\
	 y&= y_{\text{odo}}  \omega_{\text{odo}} + y_{\text{Wi-Fi}}   \omega_{\text{Wi-Fi}}. 
\end{align}
Here, $\omega_{\text{odo}}$ and $\omega_{\text{Wi-Fi}}$ represent the weighing ratios that are used to combine the odometry and Wi-Fi localization results, respectively, $(x_{\text{Wi-Fi}}, y_{\text{Wi-Fi}})$ and $(x_{\text{odo}}, y_{\text{odo}})$ are the localization values from Wi-Fi and odometry approaches respectively, and $(x,y)$ represent the calculated value of current coordinates for the robot. 

The overall algorithm for fusing the data from Wi-Fi and odometry localization are presented in Algorithm~\ref{al_main}. It is important to note that, here, we define any measured Wi-Fi coordinate $(x_{\text{Wi-Fi}}, y_{\text{Wi-Fi}})$ that differs from $(x_{\text{odo}}, y_{\text{odo}})$ by $\xi$ percentage point to be an outlier. Based on our experimental results, we have set $\xi=10\%$ throughout the rest of this paper. 

As outlined in Algorithm~\ref{al_main}, we use the dynamic weight-allocation algorithm to further reduce the absolute error in localization. The implications of this algorithm would be to dynamically change the ratio of the weights given to each localization method. Algorithm~\ref{al_dynamic} presents the approach for selecting $\omega_{\text{odo}}$ and $\omega_{\text{Wi-Fi}}$ dynamically. 



\begin{algorithm}
\caption{Dynamic Weight-Allocation Algorithm}
\label{al_dynamic}
\begin{algorithmic} 
\REQUIRE $\omega_{\text{odo}}+\omega_{\text{Wi-Fi}}=1$
\IF{$\text{odoCounter} \geq \epsilon$ and ($x_{\text{Wi-Fi}}$, $y_{\text{Wi-Fi}}$) is not an outlier}
\STATE Increase $\omega_{\text{Wi-Fi}}$ and reduce  $\omega_{\text{odo}}$
\STATE odoCounter=0
\ELSIF {($x_{\text{Wi-Fi}}$, $y_{\text{Wi-Fi}}$) is not an outlier}
\STATE $\omega_{\text{Wi-Fi}}=0.5$ and $\omega_{\text{odo}}=0.5$
\STATE odoCounter=0
\ELSE
\STATE $\omega_{\text{Wi-Fi}}=0$ and $\omega_{\text{odo}}=1$
\STATE odoCounter++
\ENDIF
\end{algorithmic}
\end{algorithm}

When the Wi-Fi localization value is deemed an outlier, the odometry credibility-counter $\--$ represented by $odoCounter$ in Algorithm~\ref{al_dynamic}$\--$increases as odometry localization values are being solely used to find the position of the robot. An $odoCounter$ larger than a set value, $\epsilon$,  denotes a low credibility in odometry positioning data due to the accumulated error. This ensures that in the next position, where the Wi-Fi localization does not give an outlier, the position of the robot is determined by giving more weight to the Wi-Fi positioning value. In contrast, a low odometry credibility -counter value denotes that the odometry measurement is reliable. As represented in Algorithm 2, when the odometry value is reliable and the Wi-Fi value is not flagged as an outlier, the weights of odometry and Wi-Fi measurements are set to $\omega_{\text{Wi-Fi}}=0.5$ and $\omega_{\text{odo}}=0.5$. The final position of the robot is then calculated as illustrated in Algorithm 1. 


%



\section{Experimental Results}\label{sec_exp}
\begin{figure}[t]
	\begin{center}
		\subfigure []
		{ \includegraphics[width=0.47\linewidth,height=5cm]{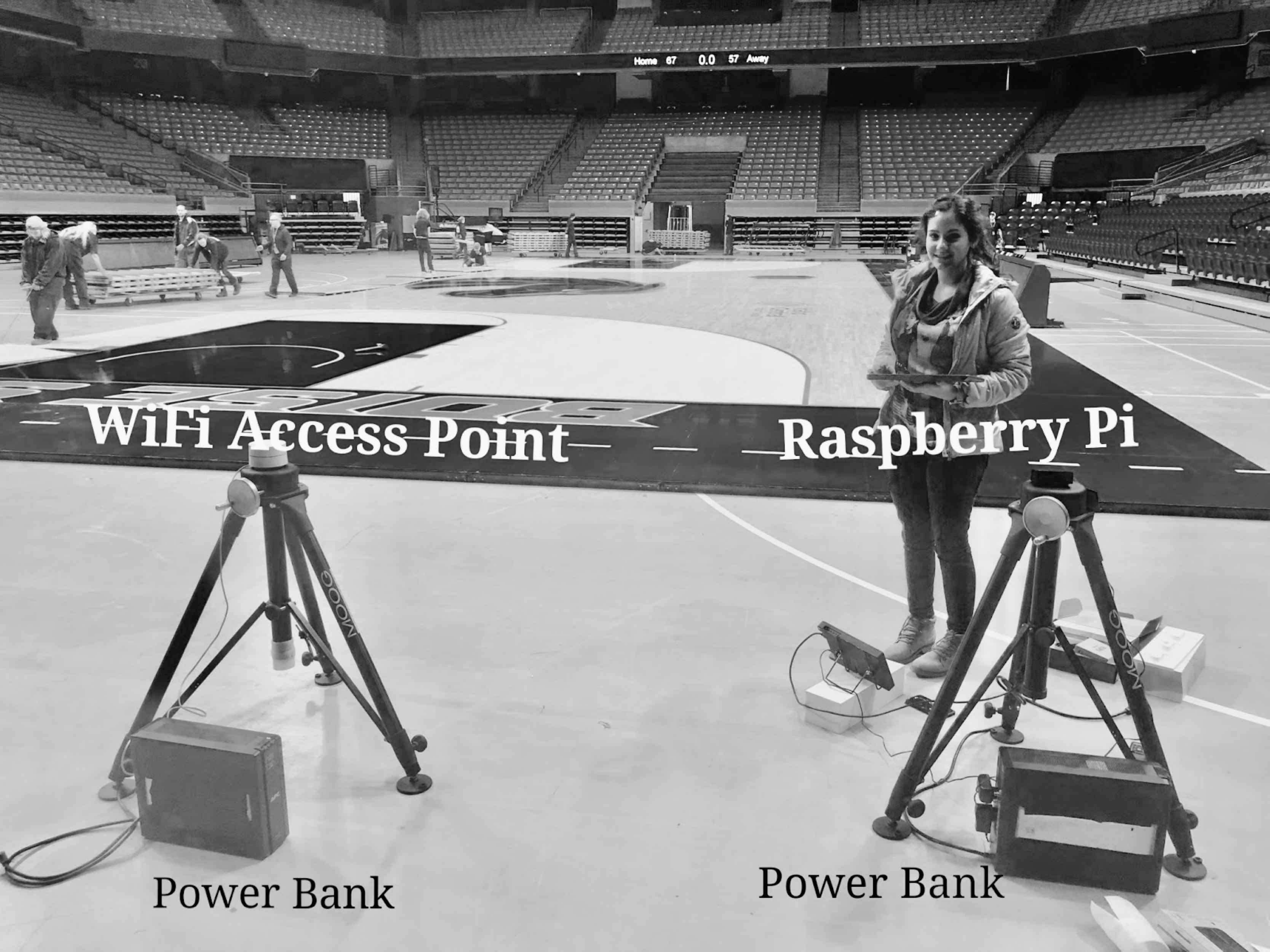}\label{fig.TBA1}}
		\subfigure []
		{\includegraphics[width=0.47\linewidth,height=5cm]{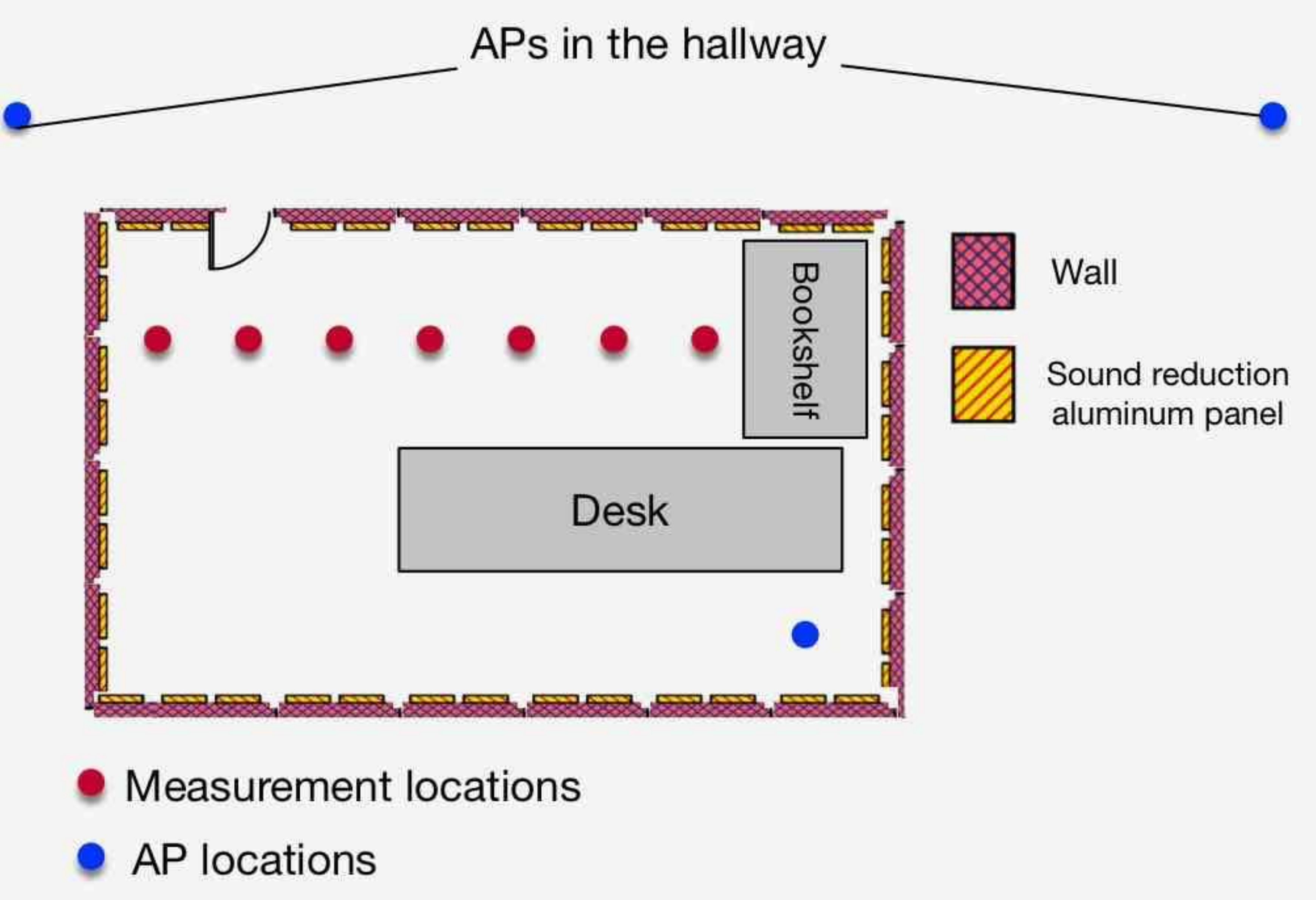}\label{fig.OFFICE}}
	\end{center}
	\caption{(a) Photo of the measurement setup. (b) Floor map of the office with the measurement locations for the robot.}
\end{figure}

\begin{figure}[t]
	\begin{center}
		\subfigure []
		{ \includegraphics[width=0.47\linewidth,height=5cm]{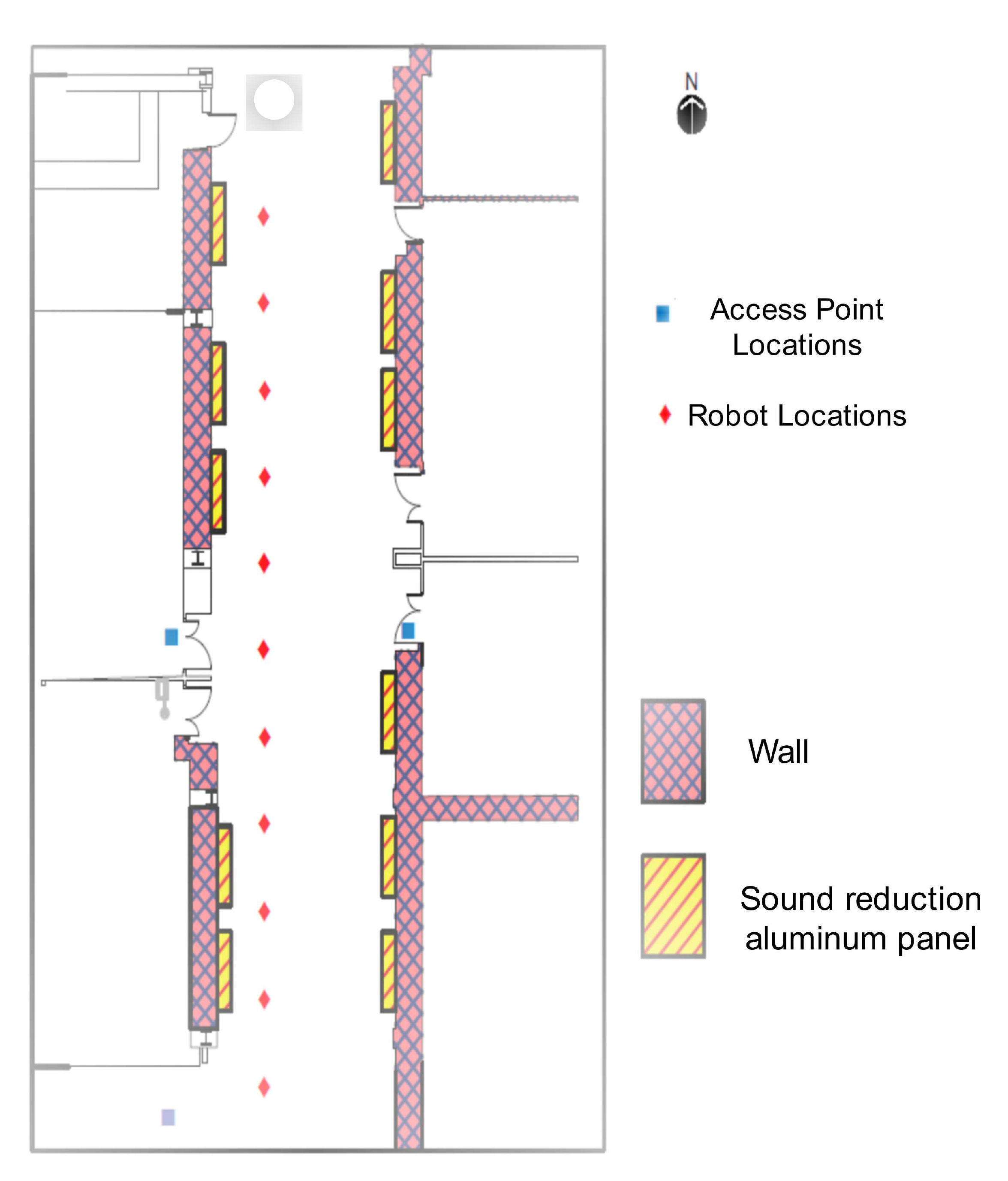}\label{fig.MEC}}
		\subfigure []
		{\includegraphics[width=0.47\linewidth,height=5cm]{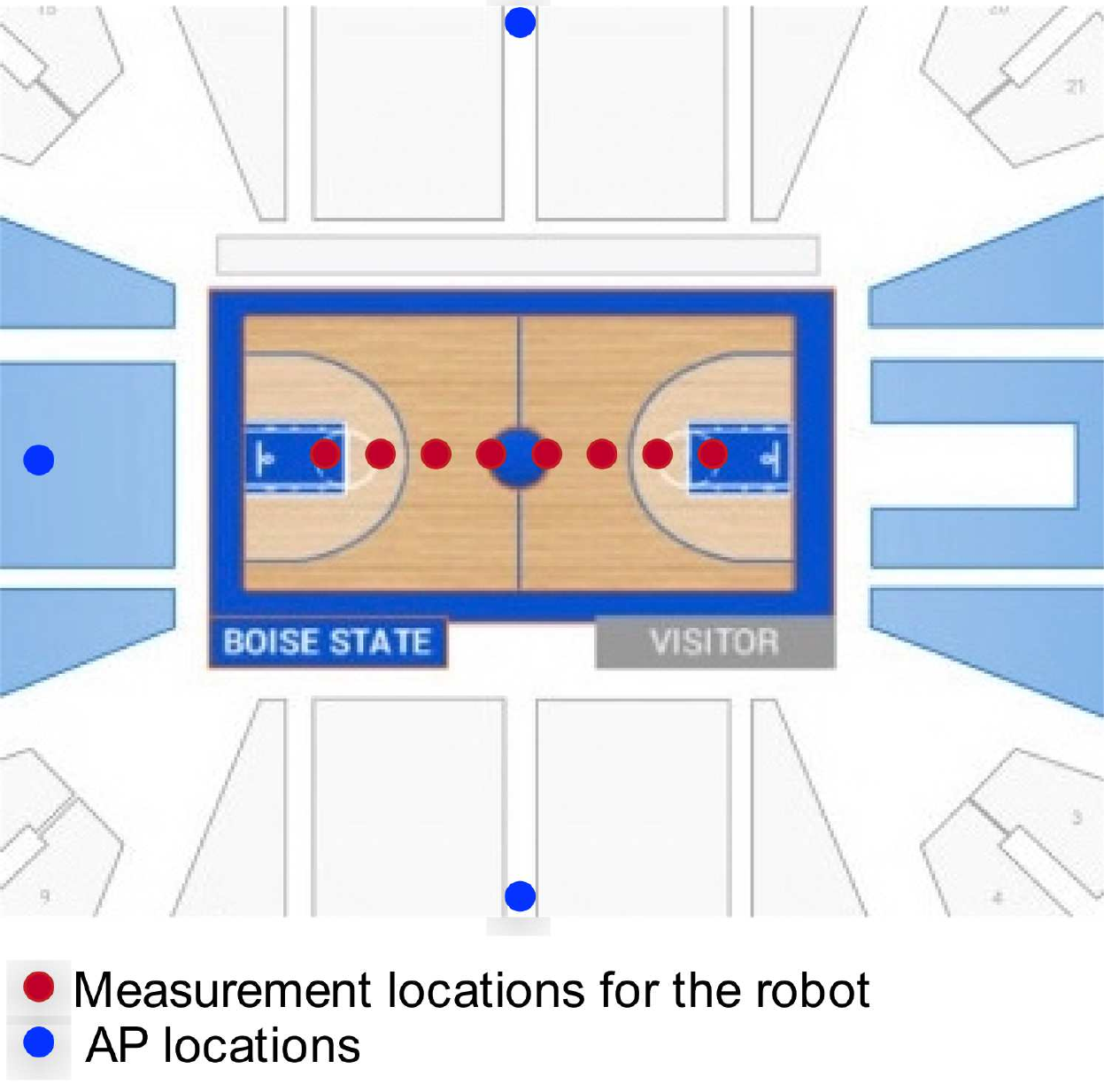}\label{fig.TBA}}
	\end{center}
	\caption{(a) Floor map of the Micron Engineering Building with the measurement locations for the robot. (b) Floor map of the Taco Bell Arena with the measurement locations for the robot.}
\end{figure}

In this section, we present experimental results using the mobile robot presented in Fig.~\ref{fig_robot}. The tests are carried out in various environments which includes an office space, the Micron Engineering Center (MEC), and Taco Bell Arena (TBA) as depicted in Figs.~\ref{fig.OFFICE},~\ref{fig.MEC}, and,~\ref{fig.TBA}, respectively. The parameters of the wireless channel in \eqref{eq1}, are set to $n=2.27$, $n=2.13$, and $n=1.71$, for the office space, the MEC, and the TBA, respectively. These parameters were determined through a measurement campaign that was carried out using a $2.4$ GHz AP and the Raspberry Pi that is installed on the robot as depicted in Fig.~\ref{fig.TBA1}. The dynamic weight-allocation algorithm has been initialized with $\omega_{odo}=0.5$ and $\omega_{Wi-Fi}=0.5$. To find the localization error, the absolute value of the difference between the measured coordinates of the robot and its true ground location are calculated. The robot traverses along the path shown in each environment, where along the way its true and measured coordinates were recorded at different points and used to obtain the error. Each experiment was repeated a number of trials to see how the error evolves. 

\begin{figure}
\begin{center}
    \scalebox{0.46}{\includegraphics {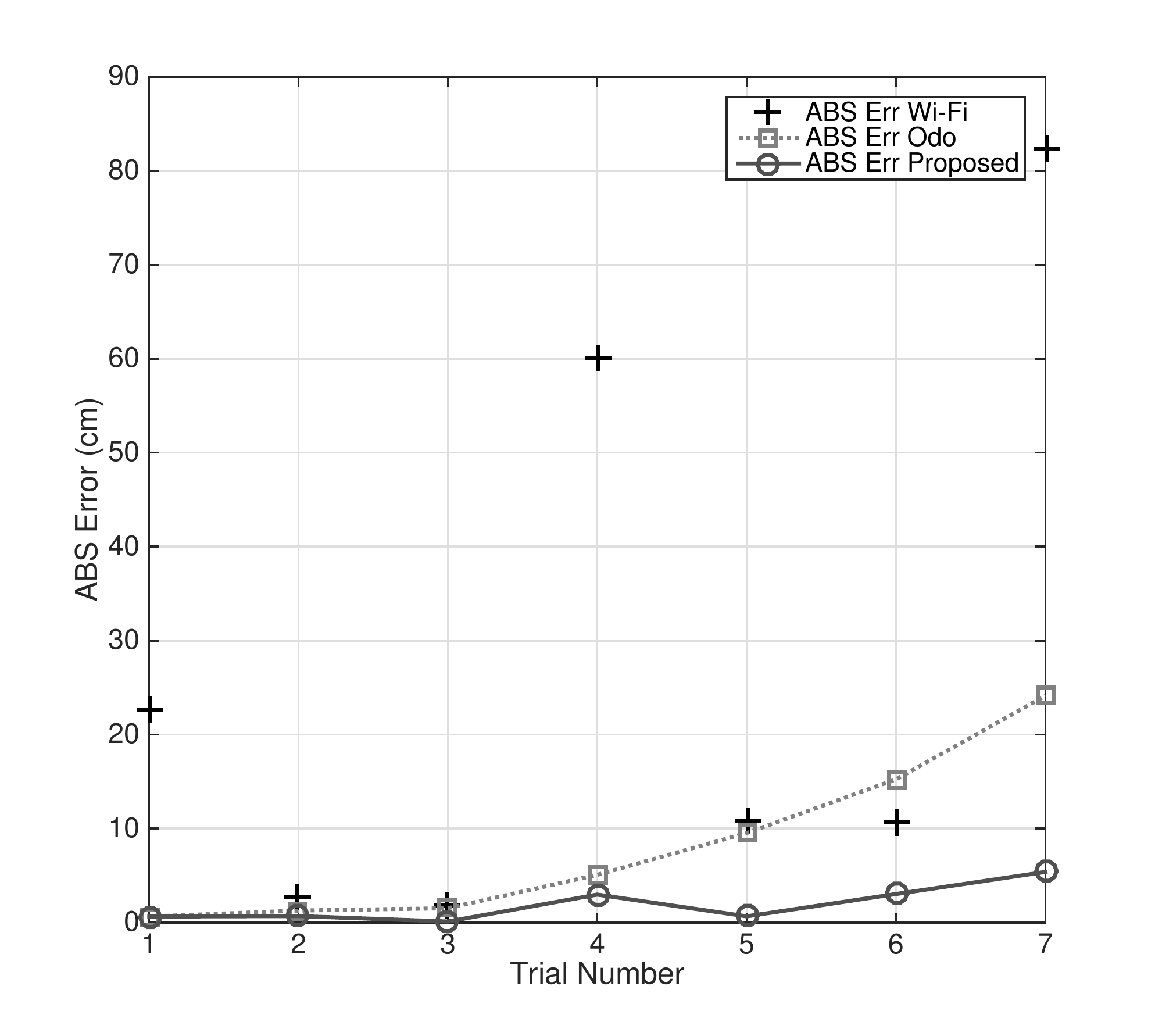}}
    \caption{Office measurement results. Absolute error in cm of odometry, Wi-Fi, and proposed method vs. number of trials.}
    \label{fig_OFFICE-RES}
\end{center}
\end{figure}

\begin{figure*}[t]
	\begin{center}
		\subfigure []
		{\hspace{-30pt}\includegraphics[width=0.50\linewidth,height=8.0cm]{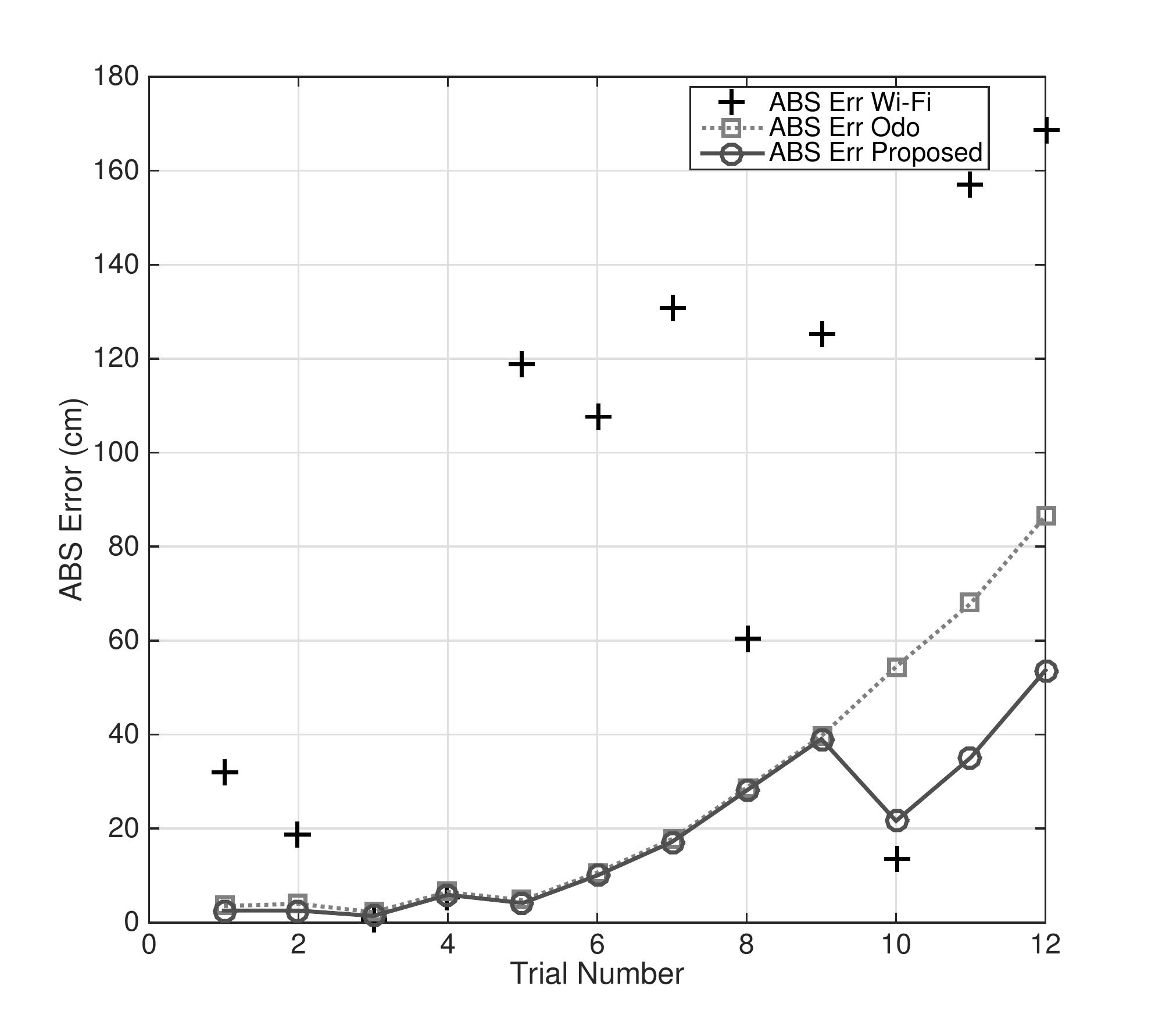}\label{fig_MEC-RES}}
		\subfigure []
		{\includegraphics[width=0.50\linewidth,height=8.0cm]{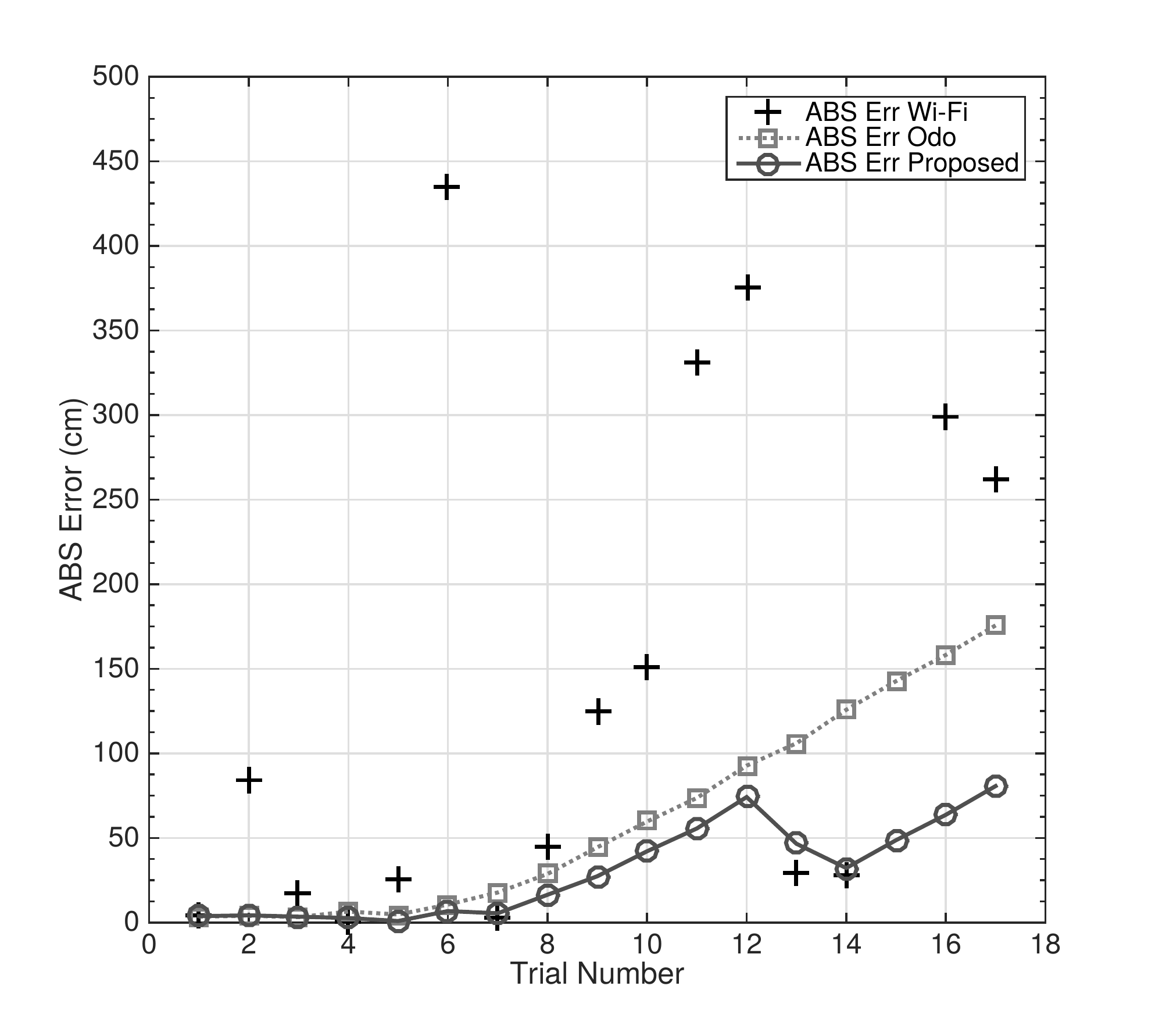}\label{fig_TBA-RES}}
	\end{center}
	\caption{(a) Micron Engineering Center measurement results. Absolute error in cm of odometry, Wi-Fi, and proposed method vs number of trials. (b) Taco Bell Arena measurement results. Absolute error in cm of odometry, Wi-Fi, and proposed method vs trial numbers.}
\end{figure*}

%

Fig.~\ref{fig_OFFICE-RES} presents the results for the measurements that were carried out inside the office environment. As anticipated, the Wi-Fi values tend to have large errors at various locations due to the shadowing factor that was described in Section~\ref{section_system_model}. At different locations, however, Wi-Fi can provide accurate localization values since there exists a direct LoS link to the receiver from the access points. As the result in Fig.~\ref{fig_OFFICE-RES} indicates, the odometry measurement values are accurate for the first few trials, but the error in odometry tends to grow with more trials. This shows the memory and relative nature of odometry measurement values. This also motivates the dynamic weight-allocation algorithm that we have proposed in Section~\ref{sec_ALG} of this paper. Finally, the results in Fig.~\ref{fig_OFFICE-RES} clearly indicate the advantage of the proposed algorithm in accurately localizing the robot. Using the proposed algorithm in $7$ trials at various points that were selected for specific measurement we see a maximum error of $5$ cms. This compared to a maximum error of $24$ cm or $82$ cm when odometry or Wi-Fi, respectively, are solely used.

Fig.~\ref{fig_MEC-RES} presents the measurements results in the MEC building of Boise State University. Here, we selected a larger number of trials. Again, we note that the proposed algorithm can outperform both Wi-Fi and odometry when used on their own. This, again, indicates that the proposed algorithm can successfully determine the outliers in Wi-Fi measurement results by fusing the localization information from odometry with that of Wi-Fi. As the results in Fig.~\ref{fig_MEC-RES} indicate, Wi-Fi resulted in a maximum error of $86$ cms, odometry resulted in a maximum error of $168$ cms, while the proposed algorithm resulted in a maximum error of $53$ cms in 12 trials runs on the path. Given the larger number of trials the overall errors tend to be larger for odometry and the proposed algorithm. 

\begin{table}
	\renewcommand{\arraystretch}{1.2}
	\caption{\bfseries  Comparison of results to that of [2]}
	\label{tab_comparison}
	\begin{center}
\begin{tabular}{ |c||c|c|c| } 
 \hline
 & Mean (cm) & Min (cm)& Max (cm)\\ 
 \hline
 \hline
Localization Error proposed & 30 & 4 & 81\\ 
\hline
Localization Error \cite{biswas2010wifi} & 70 & 20 & 160\\ 
 \hline
\end{tabular}
\end{center}     
\vspace{-10pt}                                                              
\end{table}

Finally, Fig.~\ref{fig_TBA-RES}, presents the experimental results in the Taco Bell Arena at Boise State University. Again, we see similar results as depicted in Figs.~\ref{fig_OFFICE-RES} and~\ref{fig_MEC-RES}. More specifically, Wi-Fi shows a maximum error of $452$ cms, while odometry shows a maximum error of $175$ cms when they are used without sensor fusion. Using the proposed approach, however, we see an overall improvement as the maximum error is less than halved to only $80$ cms. 18 trials runs were used in this environment.


\subsection{Comparison with Prior Work}
In Table~\ref{tab_comparison}, we compare our results to that of~\cite{biswas2010wifi}. It is important to note that a direct comparison between two different works may have discrepancies given that different experimental environments and robots are used to conduct each experiment. Comparisons between our results and other hybrid localization methods would have been ideal, however, this was not possible due to the lack of prior work regarding the combination of a wireless localization method to that of another. As the results in Table~\ref{tab_comparison} indicate, the proposed algorithm performs as well if not better than that of~\cite{biswas2010wifi}, while requiring significantly much lower overhead and complexity. The latter is true as the proposed algorithm does not require mapping the signature of Wi-Fi signal strengths at various points in an indoor environment. In fact, the only prior information we need are the location of the three APs that are used for Wi-Fi localization.

\vspace*{+6pt}
\section{Conclusion and Future Directions}
The findings in this paper indicate the effectiveness of combining Wi-Fi trilateration with odometry data. Furthermore, this paper proposes a reliable dynamic weight-allocation algorithm that fuses the sensory information from odometry and Wi-Fi localization to increase the overall efficiency and accuracy of robot localization in indoor environments. Our extensive experimental results indicate that the proposed method can more than halve the localization error of either odometry or Wi-Fi, when applied individually. Additionally, a comparison with prior work based on Wi-Fi fingerprinting shows that the proposed algorithm can match and even outperform these algorithms while significantly reducing the overhead associated with Wi-Fi localization. 

In terms of future research directions, the accuracy of odometry can be significantly increased by using a gyroscope. Additionally, this research may be applied to smartphone localization through the use of integrated mobile pedometers and gyroscopes. This also provides the dynamic weight-allocation algorithm with a more quantitative way of allocating weights to odometry measurements. Moreover, the data from a gyroscope can be used to develop a more a quantitative fusion approach, e.g. using the Bayesian framework to obtain the weights for data fusion in Section~\ref{sec_ALG}.

\section{Acknowledgment}

 We thank the NASA University Leadership Initiative Grant for funding this project. Additionally, we would like to thank the management at Taco Bell Arena, especially Ron Janeczko for giving us permission to perform our measurements at the venue.

\bibliographystyle{IEEEtran}
\bibliography{coherent_bibliography}

\begin{thebibliography}{10}
\providecommand{\url}[1]{#1}
\csname url@samestyle\endcsname
\providecommand{\newblock}{\relax}
\providecommand{\bibinfo}[2]{#2}
\providecommand{\BIBentrySTDinterwordspacing}{\spaceskip=0pt\relax}
\providecommand{\BIBentryALTinterwordstretchfactor}{4}
\providecommand{\BIBentryALTinterwordspacing}{\spaceskip=\fontdimen2\font plus
\BIBentryALTinterwordstretchfactor\fontdimen3\font minus
  \fontdimen4\font\relax}
\providecommand{\BIBforeignlanguage}[2]{{%
\expandafter\ifx\csname l@#1\endcsname\relax
\typeout{** WARNING: IEEEtran.bst: No hyphenation pattern has been}%
\typeout{** loaded for the language `#1'. Using the pattern for}%
\typeout{** the default language instead.}%
\else
\language=\csname l@#1\endcsname
\fi
#2}}
\providecommand{\BIBdecl}{\relax}
\BIBdecl

\bibitem{borenstein1997mobile}
J.~Borenstein, H.~R. Everett, L.~Feng, and D.~Wehe, ``Mobile robot positioning:
  Sensors and techniques,'' \emph{Journal Robotic Sys.}, vol.~14, no.~4, pp.
  231--249, 1997.

\bibitem{biswas2010wifi}
J.~Biswas and M.~Veloso, ``{WiFi} localization and navigation for autonomous
  indoor mobile robots,'' in \emph{Proc. IEEE int. conf. robotics and
  automation}, 2010, pp. 4379--4384.

\bibitem{wang2018iot}
D.~Wang, D.~Chen, B.~Song, N.~Guizani, X.~Yu, and X.~Du, ``From {IoT} to {5G
  I-IoT}: The next generation iot-based intelligent algorithms and {5G}
  technologies,'' \emph{IEEE Commun. Magazine}, vol.~56, no.~10, pp. 114--120,
  2018.

\bibitem{mehrpouyan2014improving}
H.~Mehrpouyan, M.~R. Khanzadi, M.~Matthaiou, A.~M. Sayeed, R.~Schober, and
  Y.~Hua, ``Improving bandwidth efficiency in e-band communication systems,''
  \emph{IEEE Commun. Magazine}, vol.~52, no.~3, pp. 121--128, 2014.

\bibitem{honkavirta2009comparative}
V.~Honkavirta, T.~Perala, S.~Ali-Loytty, and R.~Pich{\'e}, ``A comparative
  survey of {WLAN} location fingerprinting methods,'' in \emph{Proc. IEEE 6th
  workshop on positioning, navigation and communication}, 2009, pp. 243--251.

\bibitem{maxim2008trilateration}
P.~M. Maxim, S.~Hettiarachchi, W.~M. Spears, D.~F. Spears, J.~C. Hamann,
  T.~Kunkel, and C.~Speiser, ``Trilateration localization for multi-robot
  teams.'' in \emph{Proc. Inte. Conf. Informatics in Control, Automation and
  Robotics}, 2008, pp. 301--307.

\bibitem{koski2010positioning}
L.~Koski, R.~Pich{\'e}, V.~Kaseva, S.~Ali-L{\"o}ytty, and
  M.~H{\"a}nnik{\"a}inen, ``Positioning with coverage area estimates generated
  from location fingerprints,'' in \emph{Proc. IEEE 7th Workshop on
  Positioning, Navigation and Commun.}, 2010, pp. 99--106.

\bibitem{koski2010indoor}
L.~Koski, T.~Per{\"a}l{\"a}, and R.~Pich{\'e}, ``Indoor positioning using
  {WLAN} coverage area estimates,'' in \emph{Proc. IEEE Int. Conf. Indoor
  Positioning and Indoor Navigation}, 2010, pp. 1--7.

\bibitem{addesso2010adaptive}
P.~Addesso, L.~Bruno, and R.~Restaino, ``Adaptive localization techniques in
  {WiFi} environments,'' in \emph{Proc. IEEE 5th Int. Symp. on Wireless
  Pervasive Computing}, 2010, pp. 289--294.

\bibitem{wu2017mitigating}
C.~Wu, Z.~Yang, Z.~Zhou, Y.~Liu, and M.~Liu, ``Mitigating large errors in
  {WiFi}-based indoor localization for smartphones,'' \emph{IEEE Trans Vehc.
  Technology}, vol.~66, no.~7, pp. 6246--6257, 2017.

\bibitem{tao2018novel}
Y.~Tao and L.~Zhao, ``A novel system for {WiFi} radio map automatic adaptation
  and indoor positioning,'' \emph{IEEE Trans. Vehc. Technology}, vol.~67,
  no.~11, pp. 10\,683--10\,692, 2018.

\bibitem{guo2018accurate}
X.~Guo, L.~Li, N.~Ansari, and B.~Liao, ``Accurate wifi localization by fusing a
  group of fingerprints via a global fusion profile,'' \emph{IEEE Trans. Vehc.
  Technology}, vol.~67, no.~8, pp. 7314--7325, 2018.

\bibitem{sun2018augmentation}
W.~Sun, M.~Xue, H.~Yu, H.~Tang, and A.~Lin, ``Augmentation of fingerprints for
  indoor {WiFi} localization based on {Gaussian} process regression,''
  \emph{IEEE Trans. Vehc. Technology}, vol.~67, no.~11, pp. 10\,896--10\,905,
  2018.

\bibitem{wang2017csi}
X.~Wang, L.~Gao, S.~Mao, and S.~Pandey, ``{CSI-based} fingerprinting for indoor
  localization: {A} deep learning approach,'' \emph{IEEE Trans. Vehc.
  Technology}, vol.~66, no.~1, pp. 763--776, 2017.

\bibitem{rappaport2014millimeter}
T.~S. Rappaport, R.~W. Heath~Jr, R.~C. Daniels, and J.~N. Murdock,
  \emph{Millimeter wave wireless communications}.\hskip 1em plus 0.5em minus
  0.4em\relax Pearson Education, 2014.

\bibitem{maccartney2014omnidirectional}
G.~R. MacCartney, M.~K. Samimi, and T.~S. Rappaport, ``Omnidirectional path
  loss models in new york city at 28 {GHz} and 73 {GHz},'' in \emph{Proc. IEEE
  Int. Symposium}.\hskip 1em plus 0.5em minus 0.4em\relax IEEE, 2014, pp.
  227--231.

\bibitem{rappaport2015wideband}
T.~S. Rappaport, G.~R. MacCartney, M.~K. Samimi, and S.~Sun, ``Wideband
  millimeter-wave propagation measurements and channel models for future
  wireless communication system design,'' \emph{IEEE Trans. on Communications},
  vol.~63, no.~9, pp. 3029--3056, 2015.

\bibitem{goldsmith2005wireless}
A.~Goldsmith, \emph{Wireless communications}.\hskip 1em plus 0.5em minus
  0.4em\relax Cambridge university press, 2005.

\end{thebibliography}

\end{document}